\pdfoutput=1

\documentclass[11pt]{article}

\usepackage[]{naacl2021}

\usepackage{times}
\usepackage{latexsym}

\usepackage[T1]{fontenc}

\usepackage[utf8]{inputenc}

\usepackage{microtype}

\usepackage{boldline,multirow}

\usepackage{framed,color,xcolor}
\usepackage{multirow}
\usepackage{nicefrac}
\usepackage{threeparttable}
\usepackage{pgfplots}
\usepackage{soul}
\usepackage{microtype}
\usepackage{graphics}
\usepackage{framed,color,xcolor}
\usepackage{multirow}
\usepackage{nicefrac}
\usepackage{threeparttable}
\usepackage{todonotes}
\usepackage{multirow}
\usepackage{colortbl}
\usepackage{rotating, graphicx}
\usepackage{soul}      
\usepackage{colortbl}  
\usepackage{amsmath}   
\usepackage{amsfonts}  
\usepackage{amssymb}   
\usepackage{enumitem}
\usepackage{graphicx}
\usepackage{multirow}
\usepackage{colortbl}
\usepackage{rotating, graphicx}
\usepackage{multirow, makecell, caption}
\usepackage{booktabs}
\usepackage{url}
\usepackage[switch]{lineno}  %
\usepackage{arydshln}

\newcommand{\asnq}{ASNQ}
\newcommand{\astnq}{AS2-NQ}
\newcommand{\astask}{AS2} 
\newcommand{\tanda}{T{\sc and}A}
\newcommand{\ava}{AVA}
\newcommand{\diva}{RWS}

\title{Reference-based Weak Supervision for \\ Answer Sentence Selection using Web Data}

\author{Vivek Krishnamurthy\thanks{\hspace{1em}Work was conducted while the author was an intern at Amazon Alexa AI.} \\
  University of California, Los Angeles \\
  Los Angeles, CA, USA \\
  \texttt{vivek.k.murthy93@gmail.com} \\\And
  Thuy Vu \and Alessandro Moschitti\\
  Amazon Alexa AI \\ Manhattan Beach, CA, USA \\
  \texttt{thuyvu,amosch@amazon.com}  \\ 
}

\begin{document}
\maketitle
\begin{abstract}
Answer sentence selection ({\astask}) modeling requires annotated data, i.e., hand-labeled question-answer pairs.
We present a strategy to collect weakly supervised answers for a question based on its reference to improve {\astask} modeling.
Specifically, we introduce Reference-based Weak Supervision ({\diva}), a fully automatic large-scale data pipeline that harvests high-quality weakly-supervised answers from abundant Web data requiring only a question-reference pair as input.
We study the efficacy and robustness of {\diva} in the setting of {\tanda}, a recent state-of-the-art fine-tuning approach specialized for {\astask}.
Our experiments indicate that the produced data consistently bolsters {\tanda}.
We achieve the state of the art in terms of P@1, 90.1\%, and MAP, 92.9\%, on WikiQA.

\end{abstract}

\section{Introduction}

NLP advances have been brought to customers worldwide in many online services~\cite{DBLP:journals/corr/abs-1803-05567,kim-etal-2020-beyond}.
Such results combine efforts in both (i) advancing the state of the art in modeling and (ii) collecting more and better data that can maximizes model potentials.
The latter is the focus of this paper.

Creating datasets for QA requires expensive hand-labeling work.
We explore the possibility to reduce this cost by automatically leveraging abundant text data from the Web to collect weakly-supervised data, i.e., question-answer pairs.
Specifically, we propose the Reference-based Weak Supervision ({\diva}), a fully automatic data pipeline to harvest high quality answers from the Web.

{\diva} operates in two stages: (i) collecting answer candidates from Web documents and (ii) labeling them, i.e., assigning the correct or incorrect labels.
More specifically,
we build a large index of more than 100MM Web documents from Common Crawl's data. Given a question-reference pair, the question is used as query to retrieve a set of relevant documents from the index. Then, we extract sentences from those documents to build a large pool of answer candidates, which are finally scored by an automatic evaluator based on the provided reference.
We use {\ava} for our purposes, a recently introduced automatic evaluator for {\astask}~\cite{vu2020ava}.

The experimental results suggest that the weakly supervised data produced by {\diva} adds new 
\emph{supervision capacity}
to the original dataset, enabling trained models 
to advance the state of the art.
Specifically, we first verify that models trained only on {\diva} can approach the performance of models trained with the original clean data, just dropping 1-4\%.
We show that {\diva} complements the original data by measuring its improvement of  {\astask} models on WikiQA and TREC-QA datasets.

In a nutshell, our contributions include: (i) a large-scale data pipeline that generates labeled question-answer pairs using publicly available Web data, i.e., Common Crawl; and (ii) a large automatically labelled dataset derived from the data and labels of {\asnq}~\cite{garg2019tanda} with {\diva}.

\section{Background}

In this section we provide the background of our work.
We first describe {\astask} task formally, and then introduce {\tanda}, the current state-of-the-art model for {\astask}~\cite{garg2019tanda}.
Finally, we present {\ava} employed in our pipeline.

\subsection{Answer Sentence Selection ({\astask})}

The task of reranking answer candidates can be modeled with a classifier scoring the candidates.
Let $q$ be a question, $T_q=\{t_1, \dots, t_n\}$ be a set of answer candidates for $q$, we define $\mathcal{R}$ a ranking function that orders the candidates in $T_q$ according to a score, $p\left(q, t_i\right)$, indicating the probability of $t_i$ to be a correct answer for $q$. 
Popular methods modeling $\mathcal{R}$ include Compare-Aggregate~\cite{DBLP:journals/corr/abs-1905-12897}, inter-weighted alignment networks~\cite{shen-etal-2017-inter}, and Transformers~\cite{garg2019tanda}.

\begin{table}[t]
\small
\centering
\resizebox{\linewidth}{!}{%
\begin{tabular}{|lp{7.2cm}|}
\hline
$q$: & Where is the world second largest aquarium?\\
$r$: & Located in the Southeast Asian city-state of \emph{Singapore}, Marine Life Park contains twelve million gallons of water, making it the second-largest aquarium in the world.\\
$t$: & The Marine Life Park, situated in southern \emph{Singapore}, was the largest oceanarium in the world from 2012 to 2014, until it was surpassed by Chimelong Ocean Kingdom.
\\
\hline
\end{tabular}
}
\caption{A sample input for the automatic evaluator, which compares the semantic similarity between a reference $r$ and an answer candidate $t$, biased by $q$.}
\label{evaluator-input}
\vspace{-1.3em}
\end{table}

\subsection{{\tanda}: Fine-tuning for {\astask}}

\begin{figure}[h]
\centering
\includegraphics[width=\linewidth]{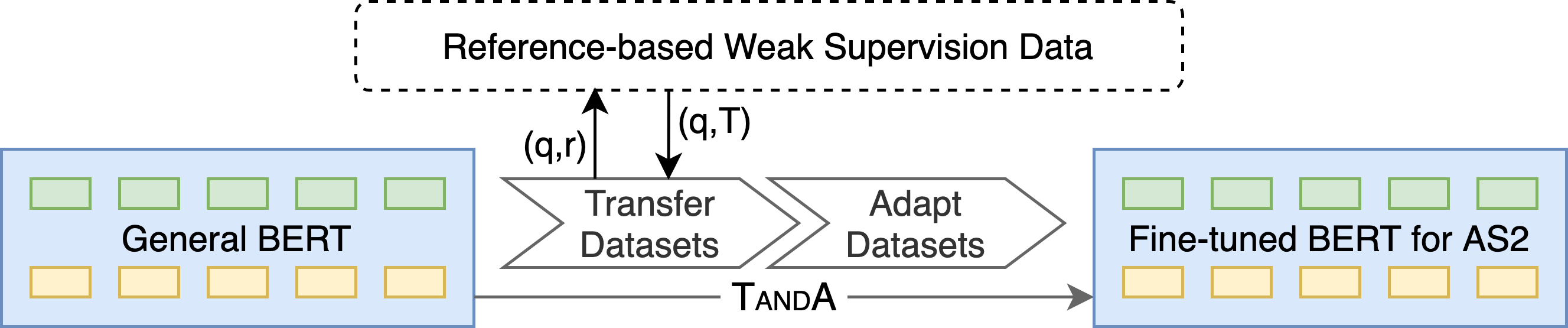}
\caption{{\diva}'s generated data applied in {\tanda}.}
\vspace{-.3em}
\label{fig:tanda_rws}
\vspace{-.5em}
\end{figure}

Fine-tuning a general pre-trained model to a target application is a recent topic of interest~\cite{gururangan-etal-2020-dont}.
Specifically, for {\astask}, \citet{garg2019tanda} introduced {\tanda}, a fine-tuning technique on using multiple datasets. {\tanda} transfers a general pre-trained Transformer model to an {\astask} model specialized on a target domain.
This approach achieved  state-of-the-art results on multiple {\astask} benchmarks.
Thus, we study and validate the robustness of {\diva} comparing with {\tanda}.
Figure~\ref{fig:tanda_rws} describes how {\diva} is integrated in {\tanda}.
In short, given a Transformer, e.g., BERT, we first fine-tune it with general datasets, including weakly supervised data, and then adapt it to the target domain using the AS2 specific data.

\subsection{Semantic Evaluator for {\astask}}
{\ava} is a recent approach to automatically measure the \emph{correctness} of an answer $t_i$ with respect to a question $q$, using a reference answer $r$.
Formally, it is modeled as a function: $\mathcal{A}\left(q, r, t_i\right) \rightarrow \{0, 1\}$, where the output is a binary correct/incorrect label. Table~\ref{evaluator-input} shows an example input for $\mathcal{A}$.

\subsection{Weakly Supervised Data Creation}
Distant supervision has gained success in creating \emph{weakly labeled data} for both relation extraction~\cite{mintz-etal-2009-distant,jiang-etal-2018-revisiting,qin-etal-2018-robust} and machine reading~\cite{joshi-etal-2017-triviaqa,kocisky-etal-2018-narrativeqa} using curated entity relation database.
Unlike others, we use abundant Web data and reference answers to create weakly label data.
We also argue that we are the first to address this research in {\astask} context.

\vspace{-.5em}
\section{Reference-based Weak Supervision}
\vspace{-.3em}

\vspace{-.3em}
\subsection{Data Generation Pipeline}
\vspace{-.3em}

\begin{figure}[t]
\centering
\includegraphics[width=\linewidth]{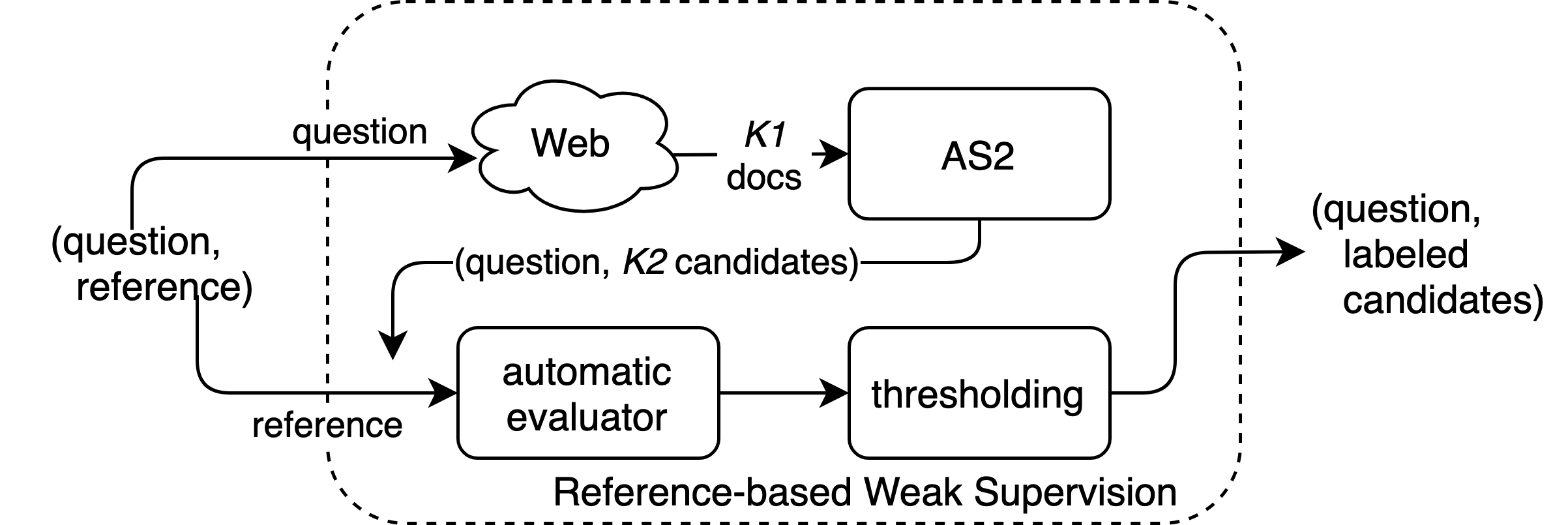}
\caption{{\diva} takes as input a (question, reference) pair and produces weakly supervised (question, answer) pairs. It consists of 4 steps: retrieval, candidate selection, automatic evaluation, and thresholding.}
\label{fig:diagram}
\vspace{-1.3em}
\end{figure}

We describe our proposed {\diva} pipeline for {\astask}.
The process starts
from
 $q$ and $r$, i.e., a valid response to $q$.
First, we retrieve top $K_1$ documents relevant to $q$ from an index of Web data.
The documents are split into sentences, which are later re-ranked by a reranker.
Second, we select the top $K_2$ sentences as candidate, $T_q=\{t_1, \dots, t_n\}$.
We create the triples of $\left(q, r, t_i\right)$  $\forall t_i \in T_q$ to be input to {\ava}, which in turns provides the scores for them.
Finally, we apply a threshold on the scores of $t_i$ to generate positive or negative labels for $t_i$.
The entire process is exemplified by Figure~\ref{fig:diagram}.
\vspace{-0.5em}

\subsection{{\ava} as an Automatic Evaluator}
{\ava} is designed to classify an answer as correct or incorrect like an {\astask} model but it exploits the semantic similarity between $t$ and  $r$, biased by $q$.
We studied multiple configurations to optimize {\ava} for our task of generating weakly supervision.
In our experiments, we use the best reported setting, which relies on a Transformer-based approach with Peer-Attention to model the interaction among $q$, $t$, and $r$ (see~\cite{vu2020ava} for a detailed technical description).
We built {\ava} using a dataset of 245 questions, each having roughly 100 annotated answers.
The number of correct and incorrect answers are 5.3K and 20.7K, respectively.
This generates approximately 500K point-wise training examples for {\ava} using the method described in~\cite{vu2020ava}. 
We verified that our training set is disjoint with respect to all 
 datasets
 studied
  in this paper to generate weakly supervised data.

\section{Experiments}
\label{sec:experiments}
We study the efficacy of {\diva} by testing its impact on {\tanda} models for {\astask}.
We first describe our experimental setup, datasets, and then
 apply
   {\diva} to {\astnq}.
We report the results of {\tanda} when {\diva}'s data is used during the transfer stage.

\subsection{Setup}

\paragraph{Large Web Index} Having the ability to query from a large index of Web documents is required in our data pipeline.
In particular, we need to retrieve a large number of documents, given a question, and we process hundreds of thousands of  questions.
As public search engines do not allow for such large-scale experimentation, we created our search engine constituted by a large index of more than $100$MM English documents, collected from 19 Common Crawl's crawls from 2013 to 2020.

\paragraph{Parameter Settings}
We employ two standard pre-trained models in our experiments: RoBERTa~\cite{DBLP:journals/corr/abs-1907-11692} and ELECTRA~\cite{Clark2020ELECTRA}.
We verify our findings on both Base and Large configurations.
We use HuggingFace’s Transformer library~\cite{Wolf2019HuggingFacesTS} and set the learning-rates to $1e\!\!-\!\!6$ and $1e\!\!-\!\!5$ for the transfer and adapt stages of {\tanda}, respectively, across all experiments.
The other hyper-parameters are set to default values.

\subsection{Datasets}
\label{sec:dataset}
We evaluated the impact of {\diva} on {\astask} using the two most popular public datasets: WikiQA and TREC-QA.
In addition, we also created {\astnq} by extending {\asnq}.
The dataset has 47\% more questions than {\asnq}, taken from the NQ dataset~\cite{47761}.
We execute {\diva} with question-reference pairs from {\astnq} and name the produced dataset {\diva} for simplicity.

\paragraph{TREC-QA} is a traditional benchmark for the AS2 task~\cite{wang-etal-2007-jeopardy}.
We use the same splits as in~\cite{garg2019tanda}.
It has 1,229, 65, and 68 questions, and 53,417,  1,117, and 1,442 question-answer pairs for train, dev., and test sets.

\paragraph{WikiQA} The dataset, introduced by ~\citet{yang2015wikiqa}, consists of questions from Bing query logs and answers extracted from a \emph{user-clicked} Wikipedia page returned by Bing.
Questions in the dataset can either have only correct answers, only incorrect answers, or mix, denoted as {\bf all$+$}, {\bf all$-$}, and {\bf clean}, respectively.
The first two are less useful in learning a reranker.
Table~\ref{table:wikiqa} reports the statistics of the dataset. 
We follow the same setting used in~\cite{garg2019tanda}: training without {\bf all$-$} and evaluate on {\bf clean}.

\begin{table}
\centering
\small
\setlength{\tabcolsep}{1.5mm}
\resizebox{0.85\linewidth}{!}{
\begin{tabular}{|r|r|r|r|r|r|r|} 
\hline
\multirow{2}{*}{~} & \multicolumn{2}{c|}{Train}                      & \multicolumn{2}{c|}{Dev}                        & \multicolumn{2}{c|}{Test}                        \\ 
\cline{2-7}
                        & \multicolumn{1}{c|}{\#Q} & \multicolumn{1}{c|}{\#A} & \multicolumn{1}{c|}{\#Q} & \multicolumn{1}{c|}{\#A} & \multicolumn{1}{c|}{\#Q} & \multicolumn{1}{c|}{\#A}  \\ 
\hline

origin                    & 2,118                   & 20,360                  & 296                    & 2,733                   & 633                    & 6,165                    \\ 
\hline
without {\bf all-}                & 873                    & 8,672                   & 126                    & 1,130                   & 243                    & 2,351                    \\ 
\hline
{\bf clean}                  & 857                    & 8,651                   & 121                    & 1,126                   & 237                    & 2,341                    \\
\hline
\end{tabular}
}
\caption{WikiQA dataset statistics: reporting the total number of questions (\#Q) and answers (\#A) for each split: Train, Dev, and Test.}
\label{table:wikiqa}
\end{table}

\paragraph{\astnq}
The transfer step in {\tanda} requires a large dataset to be effective. We built {\astnq} by extending {\asnq}.
This dataset has more than $\sim$84K questions, i.e., 27K more questions than {\asnq}, each having typically one reference answer.
The first two rows in Table~\ref{table:fromnq} show the statistics of {\asnq} and {\astnq}, respectively.

\begin{table}
\centering
\resizebox{0.85\linewidth}{!}{
\begin{tabular}{|r|r|r|r|r|} 
\hline
\multicolumn{1}{|c|}{Dataset} & \multicolumn{1}{c|}{\#Q} & \multicolumn{1}{c|}{\#A} & \multicolumn{1}{c|}{\#A$^+$} & \multicolumn{1}{c|}{\#A$^-$}  \\ 
\hline
ASNQ              & 57,242                           & 20,745,240                                   & 60,285                                     & 20,684,955                                      \\ 
\hline
{\astnq}     & 84,121                           &  27,208,065                                  & 86,756                                     &  27,121,309                                     \\ 
\hline
{\diva} & 84,089                                & 2,103,027                                            & 69,945                                     & 2,033,082                                      \\
\hline
\end{tabular}
}
\caption{Total number of questions (\#Q), answers (\#A), correct and incorrect (\#A$^+$ and \#A$^-$) of ASNQ, {\astnq}, and the weakly-supervised dataset generated from {\astnq} via our {\diva} pipeline.}
\vspace{-3mm}
\label{table:fromnq}
\end{table}

\begin{table}[h]
\centering
\resizebox{0.9\linewidth}{!}{
\begin{tabular}{|r|r|r|r|r|r|} 
\hline
\multicolumn{1}{|c|}{\multirow{2}{*}{\tanda}} & \multicolumn{1}{c|}{\multirow{2}{*}{Transfer on}} & \multicolumn{2}{c|}{WikiQA}               & \multicolumn{2}{c|}{TREC-QA}               \\ 
\cline{3-6}
\multicolumn{1}{|c|}{}                       & \multicolumn{1}{l|}{}                             & \multicolumn{1}{c|}{MAP} & \multicolumn{1}{c|}{MRR} & \multicolumn{1}{c|}{MAP} & \multicolumn{1}{c|}{MRR}  \\ 
\hline
\hline
\multirow{3}{*}{RoBERTa-Base}                & {\asnq~\shortcite{garg2019tanda}}                                              & 0.889                    & 0.901                    & {\bf 0.914}                    & {\bf 0.952}                     \\ 
\cline{2-6}
                                              & {\astnq}                                            & {\bf 0.898}                    & {\bf 0.910}                    & 0.908                    & 0.938                     \\ 
\cline{2-6}
                                             & \% diff.                                           & +1.01                   & +0.99                   & -0.66                  & -1.52                    \\ 
\hline
\clineB{1-6}{2}
\multirow{3}{*}{RoBERTa-Large}               & {\asnq~\shortcite{garg2019tanda}}                                              & 0.920                    & 0.933                    & {\bf 0.943}                    & 0.974                     \\ 
\cline{2-6}
                                              & {\astnq}                                            & {\bf 0.923}                    & {\bf 0.935}                    & 0.936                    & {\bf 0.975}                     \\
\cline{2-6}
                                             & \% diff.                                           & +0.33                   & +0.23                   & -0.73                  & +0.15                    \\

\hline
\end{tabular}}
\caption{ {\tanda}'s performance on two datasets {\asnq} and {\astnq} using RoBERTa Base and Large. {\% diff.} reports the percentage differences.}
\vspace{-.3em}
\label{table:data-compare}
\end{table}

We verified the quality of the new dataset by comparing {\tanda} models trained with {\asnq} and {\astnq}.
In particular, Table~\ref{table:data-compare} reports the results of the models when transferred on {\asnq} or  {\astnq}, measured on WikiQA and TREC-QA.
The results suggest that the end-to-end performance gain given by {\astnq} is negligible, although 47\% more data is added. This indicates that the curve amount of training data/accuracy has reached a plateau.
However, in Section~\ref{sec:diva}, we show that higher performance can still be achieved with our weakly supervised data from {\diva}.

\paragraph{{\diva}}
We apply {\diva} to {\astnq} following these steps:
First, we collect question-reference pairs from {\astnq} by using only pairs with correct answers.
We set $K_1$ and $K_2$ at 1,000 and 25, i.e., for each question, we run a query and select 1,000 relevant documents from our Elasticsearch index. This typically generates a set of 10,000 candidates. Then, we select the 25 most probable candidates using an off-the-shelf {\astask} reranker tuned on {\asnq} by~\citet{garg2019tanda}.
While a large number of questions are shared between {\asnq} and {\astnq}, the candidates from our index are disjoint.
We apply~{\ava} to label each triple, $\left(q,r,t_i\right)$, thus generating labelled pairs, $\left(q,t_i\right)$.
A pair is labeled as correct if its {\ava} score, produced by $\mathcal{A}\left(q,r,t_i\right)$, is at least 0.9, otherwise it is labeled as incorrect.

\subsection{Integrating {\diva} into {\tanda}}
\label{sec:diva}

We study the contribution of {\diva} in fine-tuning models for {\astask}.
Specifically, we compare the following transfer configurations for {\tanda}.
First, we report the baselines using (i) vanilla {\bf BERT} Base and Large models without transferring data; and (ii) {\tanda}-RoBERTa transferred with {\bf \asnq}.
We then replace {\asnq} (iii) by {\bf \astnq} and (iv) by {\bf \diva} at transfer stage, measuring the results of each transfer.
Finally, we use both datasets, {\astnq} and {\diva}, at transfer stage in the following orders: {\bf {\astnq}→{\diva}} and {\bf {\diva}→{\astnq}}.
We use precision at 1 (P@1), mean average precision (MAP), and mean reciprocal rank (MRR) as evaluation metrics.

\paragraph{General results} Table~\ref{table:all-results} shows that {\diva} used alone does not improve the baselines trained on {\asnq} or \astnq. This is intuitive as the quality of weakly supervised data is supposed to be lower than supervised data. However, when {\diva} is used as the first level of  fine-tuning (i.e., {\tanda} approach), for any dataset and any model (see model {\astnq}→*), we observed a significant improvement. In particular, when {\diva}→{\astnq} is used with RoBERTa-Large, the model establishes the new state of the art in {\astask}.

\begin{table}
\centering
\resizebox{\linewidth}{!}{
\begin{tabular}{|r|r|r|r|r|r|r|r|} 
\hline
\multicolumn{1}{|c|}{\multirow{2}{*}{PT}} & \multicolumn{1}{c|}{\multirow{2}{*}{Transfer on}} & \multicolumn{3}{c|}{WikiQA}                                                       & \multicolumn{3}{c|}{TREC-QA}                                                       \\ 
\cline{3-8}
\multicolumn{1}{|c|}{}                                   & \multicolumn{1}{c|}{}                            & \multicolumn{1}{c|}{P@1} & \multicolumn{1}{c|}{MAP} & \multicolumn{1}{c|}{MRR} & \multicolumn{1}{c|}{P@1} & \multicolumn{1}{c|}{MAP} & \multicolumn{1}{c|}{MRR}  \\ 
\hline
\hline
\clineB{1-8}{2}
\multicolumn{2}{|r|}{BERT-Base~\shortcite{garg2019tanda}}                                   & \multicolumn{1}{r|}{-} & \multicolumn{1}{r|}{0.813} & \multicolumn{1}{r|}{0.828} & \multicolumn{1}{r|}{-} & \multicolumn{1}{r|}{0.857} & \multicolumn{1}{r|}{0.937}  \\ 
\hline
\multirow{7}{*}{\rotcell[cc]{RoBERTa-Base}}  & {\asnq~\shortcite{garg2019tanda}}                                & \multicolumn{1}{r|}{-} & \multicolumn{1}{r|}{0.893} & \multicolumn{1}{r|}{0.903} & \multicolumn{1}{r|}{-} & 0.914                      & 0.952                       \\ 
\clineB{2-8}{2}
                               & {\astnq}                           & 0.852                  & 0.898                  & 0.910                  & 0.882                  & 0.908                      & 0.938                       \\ 
\cline{2-8}
                               & RWS                              & 0.716                  & 0.809                  & 0.827                  & 0.868                  & 0.878                      & 0.929                       \\ 
\cline{2-8}
                               & {\diva}→{\astnq}                              & 0.852                  & 0.897                  & 0.910                  & 0.897                  & 0.903                      & 0.945                       \\ 
                               & \% diff.                           & 0.00                    & -0.09                    & +0.02                    & +1.67                 & -0.58                    & +0.78                      \\                                
\cline{2-8}
                               & {\astnq}→{\diva}                              & {\bf 0.864}                  & {\bf 0.907}                  & {\bf 0.918}                  & {\bf 0.926}                  & {\bf 0.916}                      & {\bf 0.960}                       \\ 
                               &  \% diff.                            & +1.45                    & +0.99                    & +0.85                    & +5.00                 & +0.91                     & +2.35                      \\ 
\hline
\clineB{1-8}{2}
\multirow{7}{*}{\rotcell[cc]{Electra-Base}}  & {\asnq~\shortcite{garg2019tanda}}                             & \multicolumn{1}{r|}{-} & \multicolumn{1}{r|}{-} & \multicolumn{1}{r|}{-} & \multicolumn{1}{r|}{-} & \multicolumn{1}{r|}{-}     & \multicolumn{1}{r|}{-}      \\ 
\clineB{2-8}{2}
                               & {\astnq}                           & 0.831                  & 0.887                  & 0.896                  & 0.882                  & 0.886                      & 0.933                       \\ 
\cline{2-8}
                               & {\diva}                              & 0.712                  & 0.807                  & 0.821                  & 0.838                  & 0.827                      & 0.913                       \\ 
\cline{2-8}
                               & {\diva}→{\astnq}                              & {\bf 0.864}                  & {\bf 0.900}                  & {\bf 0.914}                  & {\bf 0.912}                  & {\bf 0.911}                      & 0.947                       \\ 
                               & \% diff.                                   & +3.96                 & +1.46                 & +1.99                 & +3.33                 & +2.74                     & +1.52                      \\ 
\cline{2-8}
                               & {\astnq}→{\diva}                              & 0.835                  & 0.890                  & 0.899                  & {\bf 0.912}                  & 0.893                      & {\bf 0.948}                       \\ 
                               &  \% diff.                                  & +0.50                 & +0.28                 & +0.42                 & +3.33                 & +0.73                     & +1.58                      \\ 
\hline
\hline
\clineB{1-8}{2}
\multicolumn{2}{|r|}{BERT-Large~\shortcite{garg2019tanda}}                                  & \multicolumn{1}{r|}{-} & \multicolumn{1}{r|}{0.836} & \multicolumn{1}{r|}{0.853} & \multicolumn{1}{r|}{-} & 0.904                      & 0.946                       \\ 
\hline
\multirow{7}{*}{\rotcell[cc]{RoBERTa-Large}} & {\asnq~\shortcite{garg2019tanda}}                             & \multicolumn{1}{r|}{-} & \multicolumn{1}{r|}{0.904} & \multicolumn{1}{r|}{0.912} & \multicolumn{1}{r|}{-} & {\bf 0.943}                      & 0.974                       \\ 
\clineB{2-8}{2.5}
                               & {\astnq}                           & 0.893                  & 0.923                & 0.935                & {\bf 0.956}                  & 0.936                      & {\bf 0.975}                       \\ 
\cline{2-8}
                               & {\diva}                              & 0.802                & 0.871                & 0.883                & 0.941                  & 0.918                      & 0.966                       \\ 
\cline{2-8}
                               & {\diva}→{\astnq}                              & {\bf 0.901}                 & {\bf 0.929}                 & {\bf 0.940}                 & 0.912                  & 0.918                      & 0.953                       \\ 
                               &    \% diff.                                & +0.92                 & +0.68                 & +0.53                 & -4.62                & -1.95                    & -2.26                     \\ 
\cline{2-8}
                               & {\astnq}→{\diva}                              & 0.889                & 0.922                & 0.935                & {\bf 0.956}                  & 0.940                      & {\bf 0.975}                       \\ 
                               &  \% diff.                                  & -0.46                & -0.16                & -0.04                & 0.00                 & +0.43                     & 0.00                      \\
\hline
\clineB{1-8}{2}
\multirow{7}{*}{\rotcell[cc]{Electra-Large}} & {\asnq~\shortcite{garg2019tanda}}                             & \multicolumn{1}{r|}{-} & \multicolumn{1}{r|}{-} & \multicolumn{1}{r|}{-} & \multicolumn{1}{r|}{-} & \multicolumn{1}{r|}{-}     & \multicolumn{1}{r|}{-}      \\ 
\clineB{2-8}{2}
                               & {\astnq}                           & 0.872                  & 0.909                  & 0.922                  & 0.941                  & 0.941                      & 0.967                       \\ 
\cline{2-8}
                               & {\diva}                              & 0.844                  & 0.894                  & 0.908                  & 0.897                  & 0.922                      & 0.949                       \\ 
\cline{2-8}
                               & {\diva}→{\astnq}                              & {\bf 0.885}                  & {\bf 0.920}                  & {\bf 0.933}                  & 0.926                  & 0.938                      & 0.957                       \\ 
                               &   \% diff.                                 & +1.42                 & +1.21                 & +1.12                 & -1.56                & -0.38                    & -1.01                     \\ 
\cline{2-8}
                               & {\astnq}→{\diva}                              & 0.885                  & 0.918                  & 0.930                  & {\bf 0.956}                  & {\bf 0.944}                      & {\bf 0.974}                       \\ 
                               &   \% diff.                                 & +1.42                 & +0.90                 & +0.84                 & +1.56                 & +0.24                     & +0.76                      \\
\hline
\end{tabular}
}
\caption{Experimental results of different {\tanda} settings on WikiQA and TREC-QA. {\bf \% diff.} indicates the performance gains (in \%) compared to the {\tanda} fine-tuned on the same {\astnq} dataset.}
\label{table:all-results}
\vspace{-2mm}
\end{table}

\vspace{-3mm}
\paragraph{WikiQA} {\diva} achieves additional performance gains when combining it with {\astnq} during the transfer steps.
In particular, we note 1\%--4\% performance gains over the {\tanda} transferred on {\astnq}.
On WikiQA, it seems better using {\diva} before {\astnq}, i.e., {\diva}→{\astnq}.

\vspace{-3mm}
\paragraph{TREC-QA} Using {\diva} during the transfer step improves the performance on TREC-QA.
While the measures are better over the baselines, i.e., using {\asnq} or {\astnq} alone, we observe a different trend during the transfer stage. Specifically, it seems more beneficial to transfer {\diva} later, i.e., {\astnq}→{\diva}.
We conjecture that this is due to the differences between WikiQA and TREC-QA.
That is, the former is very similar to {\astnq} and  {\asnq}, thus the best accuracy on WikiQA should be obtained by using {\diva} first. In contrast, TREC-QA is more general, thus it can better benefit from having {\diva}, a similar dataset, in the second step of fine-tuning.

\section{Conclusion}
We present {\diva} a fully automatic data pipeline for {\astask} that creates a large amount of weakly labeled question-answer pairs from  question-reference pairs.
This data is showed to benefit  {\astask} models.
Specifically, we recorded significant performance gains on both WikiQA and TREC-QA benchmarks.
In a nutshell, the key motivation of {\diva} is to make use of abundant Web data to find more relevant answers for a question.
We believe {\diva} can benefit other applications besides {\astask}.

\bibliography{main}

\begin{thebibliography}{18}
\expandafter\ifx\csname natexlab\endcsname\relax\def\natexlab#1{#1}\fi

\bibitem[{Clark et~al.(2020)Clark, Luong, Le, and Manning}]{Clark2020ELECTRA}
Kevin Clark, Minh-Thang Luong, Quoc~V. Le, and Christopher~D. Manning. 2020.
\newblock \href {https://openreview.net/forum?id=r1xMH1BtvB} {Electra:
  Pre-training text encoders as discriminators rather than generators}.
\newblock In \emph{International Conference on Learning Representations}.

\bibitem[{Garg et~al.(2020)Garg, Vu, and Moschitti}]{garg2019tanda}
Siddhant Garg, Thuy Vu, and Alessandro Moschitti. 2020.
\newblock \href {http://arxiv.org/abs/1911.04118} {{TANDA}: Transfer and adapt
  pre-trained transformer models for answer sentence selection}.

\bibitem[{Gururangan et~al.(2020)Gururangan, Marasovi{\'c}, Swayamdipta, Lo,
  Beltagy, Downey, and Smith}]{gururangan-etal-2020-dont}
Suchin Gururangan, Ana Marasovi{\'c}, Swabha Swayamdipta, Kyle Lo, Iz~Beltagy,
  Doug Downey, and Noah~A. Smith. 2020.
\newblock \href {https://doi.org/10.18653/v1/2020.acl-main.740} {Don{'}t stop
  pretraining: Adapt language models to domains and tasks}.
\newblock In \emph{Proceedings of the 58th Annual Meeting of the Association
  for Computational Linguistics}, pages 8342--8360, Online. Association for
  Computational Linguistics.

\bibitem[{Hassan et~al.(2018)Hassan, Aue, Chen, Chowdhary, Clark, Federmann,
  Huang, Junczys{-}Dowmunt, Lewis, Li, Liu, Liu, Luo, Menezes, Qin, Seide, Tan,
  Tian, Wu, Wu, Xia, Zhang, Zhang, and
  Zhou}]{DBLP:journals/corr/abs-1803-05567}
Hany Hassan, Anthony Aue, Chang Chen, Vishal Chowdhary, Jonathan Clark,
  Christian Federmann, Xuedong Huang, Marcin Junczys{-}Dowmunt, William Lewis,
  Mu~Li, Shujie Liu, Tie{-}Yan Liu, Renqian Luo, Arul Menezes, Tao Qin, Frank
  Seide, Xu~Tan, Fei Tian, Lijun Wu, Shuangzhi Wu, Yingce Xia, Dongdong Zhang,
  Zhirui Zhang, and Ming Zhou. 2018.
\newblock \href {http://arxiv.org/abs/1803.05567} {Achieving human parity on
  automatic chinese to english news translation}.
\newblock \emph{CoRR}, abs/1803.05567.

\bibitem[{Jiang et~al.(2018)Jiang, Liu, Lin, and
  Sui}]{jiang-etal-2018-revisiting}
Tingsong Jiang, Jing Liu, Chin-Yew Lin, and Zhifang Sui. 2018.
\newblock \href {https://www.aclweb.org/anthology/L18-1566} {Revisiting distant
  supervision for relation extraction}.
\newblock In \emph{Proceedings of the Eleventh International Conference on
  Language Resources and Evaluation ({LREC} 2018)}, Miyazaki, Japan. European
  Language Resources Association (ELRA).

\bibitem[{Joshi et~al.(2017)Joshi, Choi, Weld, and
  Zettlemoyer}]{joshi-etal-2017-triviaqa}
Mandar Joshi, Eunsol Choi, Daniel Weld, and Luke Zettlemoyer. 2017.
\newblock \href {https://doi.org/10.18653/v1/P17-1147} {{T}rivia{QA}: A large
  scale distantly supervised challenge dataset for reading comprehension}.
\newblock In \emph{Proceedings of the 55th Annual Meeting of the Association
  for Computational Linguistics (Volume 1: Long Papers)}, pages 1601--1611,
  Vancouver, Canada. Association for Computational Linguistics.

\bibitem[{Kim et~al.(2020)Kim, Eric, Gopalakrishnan, Hedayatnia, Liu, and
  Hakkani-Tur}]{kim-etal-2020-beyond}
Seokhwan Kim, Mihail Eric, Karthik Gopalakrishnan, Behnam Hedayatnia, Yang Liu,
  and Dilek Hakkani-Tur. 2020.
\newblock \href {https://www.aclweb.org/anthology/2020.sigdial-1.35} {Beyond
  domain {API}s: Task-oriented conversational modeling with unstructured
  knowledge access}.
\newblock In \emph{Proceedings of the 21th Annual Meeting of the Special
  Interest Group on Discourse and Dialogue}, pages 278--289, 1st virtual
  meeting. Association for Computational Linguistics.

\bibitem[{Ko{\v{c}}isk{\'y} et~al.(2018)Ko{\v{c}}isk{\'y}, Schwarz, Blunsom,
  Dyer, Hermann, Melis, and Grefenstette}]{kocisky-etal-2018-narrativeqa}
Tom{\'a}{\v{s}} Ko{\v{c}}isk{\'y}, Jonathan Schwarz, Phil Blunsom, Chris Dyer,
  Karl~Moritz Hermann, G{\'a}bor Melis, and Edward Grefenstette. 2018.
\newblock \href {https://doi.org/10.1162/tacl_a_00023} {The {N}arrative{QA}
  reading comprehension challenge}.
\newblock \emph{Transactions of the Association for Computational Linguistics},
  6:317--328.

\bibitem[{Kwiatkowski et~al.(2019)Kwiatkowski, Palomaki, Redfield, Collins,
  Parikh, Alberti, Epstein, Polosukhin, Kelcey, Devlin, Lee, Toutanova, Jones,
  Chang, Dai, Uszkoreit, Le, and Petrov}]{47761}
Tom Kwiatkowski, Jennimaria Palomaki, Olivia Redfield, Michael Collins, Ankur
  Parikh, Chris Alberti, Danielle Epstein, Illia Polosukhin, Matthew Kelcey,
  Jacob Devlin, Kenton Lee, Kristina~N. Toutanova, Llion Jones, Ming-Wei Chang,
  Andrew Dai, Jakob Uszkoreit, Quoc Le, and Slav Petrov. 2019.
\newblock \href
  {https://tomkwiat.users.x20web.corp.google.com/papers/natural-questions/main-1455-kwiatkowski.pdf}
  {Natural questions: a benchmark for question answering research}.
\newblock \emph{TACL}.

\bibitem[{Liu et~al.(2019)Liu, Ott, Goyal, Du, Joshi, Chen, Levy, Lewis,
  Zettlemoyer, and Stoyanov}]{DBLP:journals/corr/abs-1907-11692}
Yinhan Liu, Myle Ott, Naman Goyal, Jingfei Du, Mandar Joshi, Danqi Chen, Omer
  Levy, Mike Lewis, Luke Zettlemoyer, and Veselin Stoyanov. 2019.
\newblock \href {http://arxiv.org/abs/1907.11692} {Roberta: {A} robustly
  optimized {BERT} pretraining approach}.
\newblock \emph{CoRR}, abs/1907.11692.

\bibitem[{Mintz et~al.(2009)Mintz, Bills, Snow, and
  Jurafsky}]{mintz-etal-2009-distant}
Mike Mintz, Steven Bills, Rion Snow, and Daniel Jurafsky. 2009.
\newblock \href {https://www.aclweb.org/anthology/P09-1113} {Distant
  supervision for relation extraction without labeled data}.
\newblock In \emph{Proceedings of the Joint Conference of the 47th Annual
  Meeting of the {ACL} and the 4th International Joint Conference on Natural
  Language Processing of the {AFNLP}}, pages 1003--1011, Suntec, Singapore.
  Association for Computational Linguistics.

\bibitem[{Qin et~al.(2018)Qin, Xu, and Wang}]{qin-etal-2018-robust}
Pengda Qin, Weiran Xu, and William~Yang Wang. 2018.
\newblock \href {https://doi.org/10.18653/v1/P18-1199} {Robust distant
  supervision relation extraction via deep reinforcement learning}.
\newblock In \emph{Proceedings of the 56th Annual Meeting of the Association
  for Computational Linguistics (Volume 1: Long Papers)}, pages 2137--2147,
  Melbourne, Australia. Association for Computational Linguistics.

\bibitem[{Shen et~al.(2017)Shen, Yang, and Deng}]{shen-etal-2017-inter}
Gehui Shen, Yunlun Yang, and Zhi-Hong Deng. 2017.
\newblock \href {https://doi.org/10.18653/v1/D17-1122} {Inter-weighted
  alignment network for sentence pair modeling}.
\newblock In \emph{EMNLP'17}, pages 1179--1189, Copenhagen, Denmark.

\bibitem[{Vu and Moschitti(2021)}]{vu2020ava}
Thuy Vu and Alessandro Moschitti. 2021.
\newblock Ava: an automatic evaluation approach to question answering systems.
\newblock In \emph{NAACL}.

\bibitem[{Wang et~al.(2007)Wang, Smith, and Mitamura}]{wang-etal-2007-jeopardy}
Mengqiu Wang, Noah~A. Smith, and Teruko Mitamura. 2007.
\newblock \href {https://www.aclweb.org/anthology/D07-1003} {What is the
  {J}eopardy model? a quasi-synchronous grammar for {QA}}.
\newblock In \emph{{EMNLP}-{C}o{NLL}'07}, pages 22--32, Prague, Czech Republic.
  Association for Computational Linguistics.

\bibitem[{Wolf et~al.(2019)Wolf, Debut, Sanh, Chaumond, Delangue, Moi, Cistac,
  Rault, Louf, Funtowicz, and Brew}]{Wolf2019HuggingFacesTS}
Thomas Wolf, Lysandre Debut, Victor Sanh, Julien Chaumond, Clement Delangue,
  Anthony Moi, Pierric Cistac, Tim Rault, R'emi Louf, Morgan Funtowicz, and
  Jamie Brew. 2019.
\newblock Huggingface's transformers: State-of-the-art natural language
  processing.
\newblock \emph{ArXiv}, abs/1910.03771.

\bibitem[{Yang et~al.(2015)Yang, Yih, and Meek}]{yang2015wikiqa}
Yi~Yang, Wen-tau Yih, and Christopher Meek. 2015.
\newblock Wikiqa: A challenge dataset for open-domain question answering.
\newblock In \emph{Proceedings of the 2015 conference on empirical methods in
  natural language processing}, pages 2013--2018.

\bibitem[{Yoon et~al.(2019)Yoon, Dernoncourt, Kim, Bui, and
  Jung}]{DBLP:journals/corr/abs-1905-12897}
Seunghyun Yoon, Franck Dernoncourt, Doo~Soon Kim, Trung Bui, and Kyomin Jung.
  2019.
\newblock \href {http://arxiv.org/abs/1905.12897} {A compare-aggregate model
  with latent clustering for answer selection}.
\newblock \emph{CoRR}, abs/1905.12897.

\end{thebibliography}
\bibliographystyle{acl_natbib}

\end{document}